\pdfoutput=1

\documentclass[11pt]{article}

\usepackage[final]{acl}

\usepackage{times}
\usepackage{latexsym}
\usepackage{svg}
\usepackage{graphicx}%
\usepackage{multirow}%
\usepackage{arydshln}
\usepackage{amsmath,amssymb,amsfonts}%
\usepackage{amsthm}%
\usepackage{mathrsfs}%
\usepackage[title]{appendix}%
\usepackage{textcomp}%
\usepackage{manyfoot}%
\usepackage{booktabs}%
\usepackage{algorithm}%
\usepackage{algorithmicx}%
\usepackage{algpseudocode}%
\usepackage{listings}%
\usepackage{graphicx}
\usepackage{tcolorbox}
\usepackage{soul}
\usepackage{amssymb}
\usepackage{pifont}
\newcommand{\cmark}{\ding{51}}%
\newcommand{\xmark}{\ding{55}}%
\usepackage{appendix}
\usepackage{caption}
\usepackage{multirow}
\usepackage{makecell}
\usepackage{colortbl}
\usepackage{subcaption}
\usepackage{array} 
\usepackage{hyperref}

\usepackage[T1]{fontenc}

\usepackage[utf8]{inputenc}

\usepackage{microtype}

\usepackage{inconsolata}

\usepackage{graphicx}
\newcommand{\model}{KaLMA}
\newcommand{\visTEL}{VisTEL}
\newcommand{\data}{{\sc T}ext-{\textsc{kvqa}}}
\title{Visual Text Matters: Improving Text-KVQA with Visual Text Entity Knowledge-aware Large Multimodal Assistant}

\author{Abhirama Subramanyam Penamakuri and Anand Mishra \\
  Indian Institute of Technology Jodhpur\\
  \texttt{\{penamakuri.1,mishra\}@iitj.ac.in}\\
  \href{https://vl2g.github.io/projects/LMM4Text-KVQA/}{\textbf{{https://vl2g.github.io/projects/LMM4Text-KVQA/}}}}

\begin{document}
\maketitle
\begin{abstract}
We revisit knowledge-aware text-based visual question answering, also known as \data{} in the light of modern advancements in large multimodal models ({\sc lmm}s), and make the following contributions: (i) We propose \visTEL{} -- a principled approach to perform visual text entity linking. The proposed \visTEL{} module harnesses a state-of-the-art visual text recognition engine and the power of a large multimodal model to jointly reason using textual and visual context obtained using surrounding cues in the image to link the visual text entity to the correct knowledge base entity. (ii) We present \model{} -- knowledge-aware large multimodal assistant that augments an {\sc lmm} with knowledge associated with visual text entity in the image to arrive at an accurate answer. Further, we provide a comprehensive experimental analysis and comparison of our approach with traditional visual question answering, pre-large multimodal models, and large multimodal models, as well as prior top-performing approaches. Averaging over three splits of \data{}, our proposed approach surpasses the previous best approach by a substantial 23.3\% on an absolute scale and establishes a new state of the art. We make our implementation publicly available. 
\end{abstract}

\section{Introduction}
\begin{figure*}[!t]
\centering
  \includegraphics[width=\textwidth]{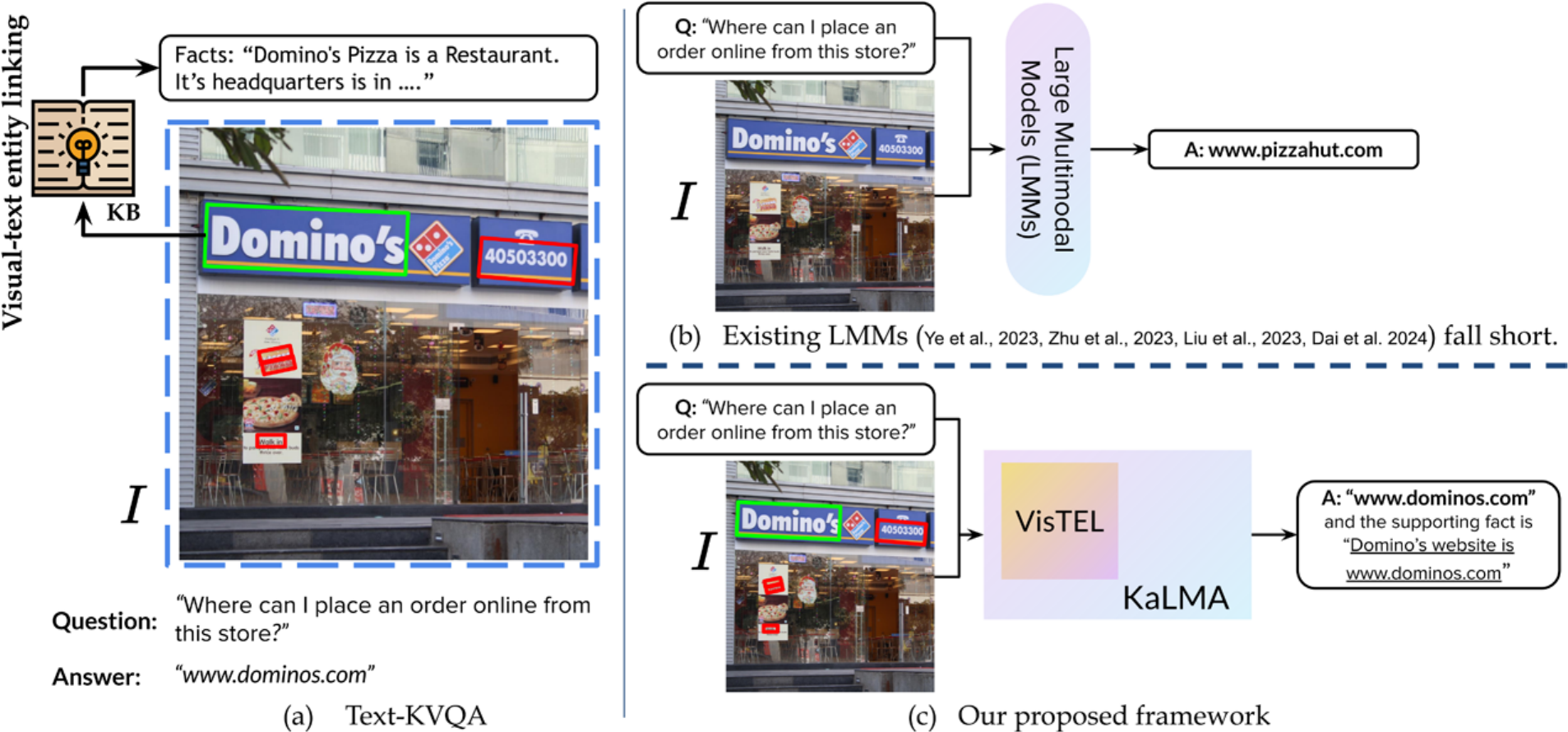}
  \caption{(a) \textbf{\data{}}~\cite{singh2019strings}: Given an image containing a named entity as visual text, e.g., ``Domino's" in this illustration, the aim is to answer the question by leveraging explicit knowledge about the visual text entity. (b) Large Multimodal Models are one obvious choice for solving such tasks today. However, they alone are insufficient as they hallucinate on visual objects. (c) We propose a novel approach -- \model{} that augments an {\sc lmm} with specialized visual text recognition and retrieved relevant knowledge obtained using visual text entity linking by proposed \visTEL{}. Our approach establishes a new state-of-the-art for this task.}
  \label{fig:goal}
\end{figure*}
In the past few years, the research community has shown significant interest in visual question answering based on text appearing in images, as evidenced by the emergence of {\sc ocr-vqa}~\cite{mishra2019ocr}, {\sc st-vqa}~\cite{biten2019scene} and {\sc t}ext{\sc vqa}~\cite{singh2019towards}. Giving another aspect to these problems by leveraging external knowledge for text-based visual question answering, ~\cite{singh2019strings} introduced a task called \data{}. The \data{} presents a unique challenge: given an image containing textual entities like business brands, book titles, or movie titles, the task is to answer questions that require external knowledge about these entities. 
Addressing \data{} involves detecting text in images, recognizing it, linking it to a knowledge base, and employing visual context and knowledge base for reasoning to provide an answer. Since the introduction of this problem, several advancements have happened in visual text understanding as well as vision and language models. In this work, we revisit \data{} by leveraging these modern advancements and propose a framework that judiciously integrates various components of contemporary architecture.

The emergence of large multimodal models ({\sc lmm}s)\footnote{We refer to both large multimodal model and large vision and language models as {\sc lmm} in this work.} represents a significant trend in the literature on vision and language~\cite{zhang2022opt,flant5,touvron2023llama,liu2023llava,zhu2023minigpt,ye2023mplug,falcon,instructgpt}. Over the past few years, many large-scale language and vision models have been developed, demonstrating exceptional performance across various tasks including, but not limited to, image captioning, visual question answering, multimodal reasoning, and visual grounding. We believe that pretrained LMMs hold great potential for addressing \data{}. These models are rich in the implicit knowledge learned by large-scale pretraining. However, despite their numerous advantages, they are not without drawbacks, notably hallucinations. This challenge becomes particularly apparent in \data{}, where precise reasoning about entities depicted in images and associated knowledge is required. Consider the following scenario where a customer, after finishing their meal at a restaurant store, takes a picture of the store signboard and enquires about a possible future online delivery, asking, `Where can I place an online order from this store?' (Figure~\ref{fig:goal}(a)).  Existing {\sc lmm}s often hallucinate over the pizza present in the image and points to the website of `Pizza Hut' instead of `Domino's' (Figure~\ref{fig:goal}(b)); whereas complementing the {\sc lmm} with an explicit visual text entity linking followed by knowledge-retrieval helps overcome hallucination (Figure~\ref{fig:goal}(c)), thereby generating an accurate answer to the given question. Our model is developed on this hypothesis. 

We address \data{} by introducing an architecture, namely \model{} -- knowledge-aware large multimodal assistant that first invokes our proposed visual text entity linker or \visTEL{} -- an {\sc lmm}-architecture that links visual text entities to the associated knowledge base (Illustrated in Figure~\ref{fig:goal} (c)). Once the entities are linked to the knowledge base, the associated knowledge is retrieved and augmented to a large multimodal model to answer visual questions. 

To summarize, our contributions are as follows: (i) We revisit \data{} -- a task originally introduced by~\cite{singh2019strings} in the light of the latest advancements in large multimodal models. To this end, we benchmark latest {\sc lmm}s on \data{}. Our study highlights that {\sc lmm}s although powerful, often ignore visual text present in the images, resulting in hallucinations. 

(ii) We propose a principled approach called \visTEL{} for linking visual text entities that appear in images to a knowledge base. \visTEL{} is an {\sc lmm}-based architecture that leverages the surrounding OCR-extracted texts obtained using a specialized text recognition module and the visual context within the image to perform highly accurate entity linking for visual text entities. (iii) We introduce \model{} -- a Knowledge-aware Large Multimodal Assistant, which enhances a large-multimodal model, specifically LLaVA~\cite{liu2023llava} by integrating retrieved knowledge from our proposed \visTEL{}. This augmentation facilitates robust vision and language reasoning, thereby enabling superior knowledge-aware text-based visual question answering. (iv) We conduct extensive experiments and ablation to show the superior performance of our proposed framework over competitive approaches and state of the art. We provide several exciting insights about our design choice, attribution ability of \model{}, and addressing hallucination issues of {\sc lmm}s. Our proposed approach advances state of the art on \data{} by 18.2\% on scene, 19.6\% on book covers, and 32.2\% on movie poster splits of the dataset on an absolute scale. 

\section{Related Work}
\noindent\textbf{KVQA Tasks:} Visual Question Answering is a well-studied task~\cite{antol2015vqa,goyal2017making}. This task has been extended to scenarios that require the ability to read text within images, leading to the development benchmarks such as {\sc st-vqa}~\cite{biten2019scene,biten2019icdar}, {\sc t}ext{\sc vqa}~\cite{singh2019towards}, {\sc d}oc{\sc vqa}~\cite{mathew2021docvqa}, and {\sc ocr-vqa}~\cite{mishra2019ocr}. While these benchmarks were successful in their intent of integrating reading and reasoning abilities in VQA, they are often restricted to reasoning around what is visually apparent. To address this gap and encourage models to perform reasoning beyond visually apparent facts, \cite{singh2019strings} introduced knowledge-aware Text-based VQA task. Distinctively different from other knowledge-aware visual question answering tasks such as {\sc kb-vqa}~\cite{wangkbvqa}, {\sc fvqa}~\cite{wang2017fvqa}, {\sc kvqa}~\cite{shah2019kvqa}, {\sc ok-vqa}~\cite{marino2019okvqa}, and Infoseek~\cite{infoseek}, \data{} deals with reasoning over visual text entities and associated knowledge to arrive at answer.

\noindent\textbf{Methods Prior to Large Multimodal Models:}
Early methods to solve knowledge-aware VQA tasks focus on leveraging knowledge in the form of triplets~\cite{narasimhan2018out,narasimhan2018straight,wu2016ask}, or sub-knowledge-graph~\cite{zhang2018variational,singh2019strings} or memory facts~\cite{weston2015memory}. Later, transformer architectures~\cite{vaswani2017attention} owing to their ability to encode intrinsic knowledge using large-scale pretraining, have become defacto for addressing {\sc kvqa}. 

Inspired by the hybrid models, e.g.~\cite{rag,realm} where intrinsic knowledge of transformer architectures is complemented with explicit external knowledge; researchers proposed hybrid methods such as {\sc c}oncept{\sc bert}~\cite{garderes-etal-2020-conceptbert}, {\sc krisp}~\cite{marino2021krisp}, and {\sc reveal}~\cite{hu2023reveal} which augment the multimodal transformers with explicitly retrieved external knowledge. 

\noindent\textbf{Emergence of Large Multimodal Models:}
The early success of large-scale pretraining on the downstream tasks demonstrated by the foundation models, e.g., {\sc bert}~\cite{devlin-etal-2019-bert} and {\sc gpt}~\cite{gpt2} paved the way for the researchers to scale the model and the data used for pretraining. {\sc gpt-3}~\cite{gpt3} is an early large language model ({\sc llm}) demonstrating reliable performance on many downstream tasks. Following this, several {\sc llm} variants~\cite{zhang2022opt,flant5,workshop2022bloom,falcon,touvron2023llama} have been introduced. Researchers adopted these {\sc llm}s to vision-language research, with the key idea being aligning the visual information with the linguistic information of the {\sc llm}s to come up with large multimodal models ({\sc lmm}s)~\cite{frozen,falcon,blip2,zhu2023minigpt,ye2023mplug,liu2023llava}. Recently, {\sc lmm}s have become first-hand solutions for many downstream vision-language tasks, making them an obvious choice to solve \data{}. Authors in~\cite{pica,mmreasoner} prompt the {\sc llm}s with visual information via dense captions, object tags, object-level bounding box coordinates, and OCR tags. 
These methods rely heavily on the implicit knowledge learned by these {\sc llm}s. Further, {\sc kat}~\cite{gui2022kat} improves upon such methods by augmenting external knowledge via retriever before prompting the {\sc llm}. However, it ignores the explicit visual information, which {\sc revive}~\cite{lin2022revive} aims to fix. Although these methods show significant success, they have limitations such as hallucination and ignoring visual texts for reasoning. We aim to fill these gaps by proposing a novel solution for \data{}.

\noindent\textbf{Visual Entity Linking:} 
Entity linking has traditionally been a well-established focus area within the NLP community~\cite{10.5555/1214993}. In contrast, the problem of visual entity linking has only garnered attention in the last decade~\cite{HuLCKJLTC23,SunFG0C22,shah2019kvqa}.~\cite{SunFG0C22} have proposed a novel dataset and benchmark for visual named entity linking.~\cite{shah2019kvqa} drew attention to the need for visual entity linking for addressing knowledge-based visual question answering. Open-domain Visual Entity Recognition has also been studied in the literature~\cite{HuLCKJLTC23,caron2024generative,xiao2024grounding}. However, most of these works have focused on linking entities such as persons, landmarks, and other named entities, while neglecting visual text such as business brand names and movie or book titles. In this work, we address this gap by proposing a principled solution for visual text entity linking and demonstrate its utility as a precursor to \data{}.

\section{Methodology}
\label{section:method}
\begin{figure*}[!t]
\centering
  \includegraphics[width=\textwidth]{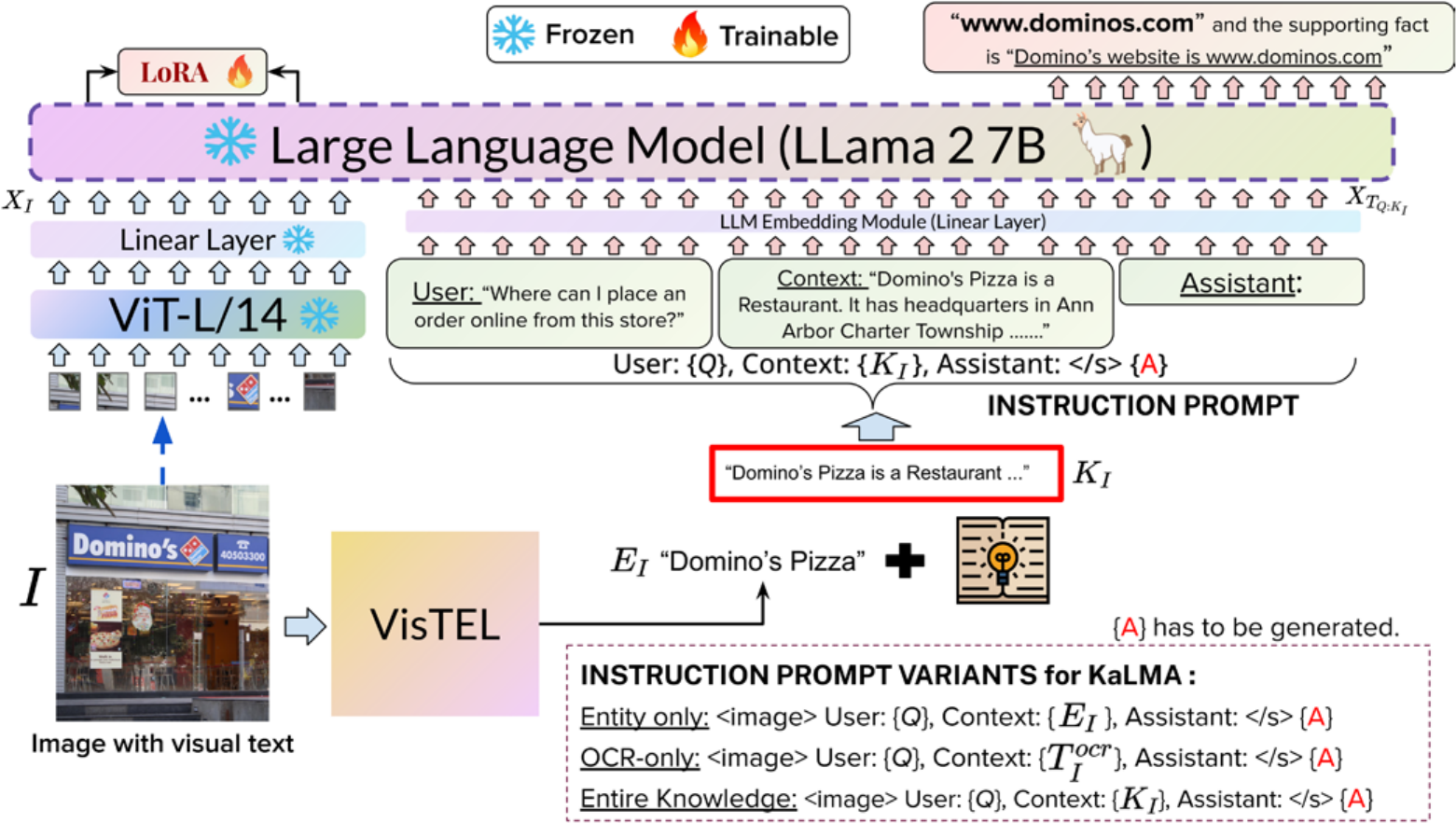}
  \caption{\textbf{Overview of our proposed framework \model{}}. We first link the visual text in the image $I$ to the entity $E_I$ using \visTEL{} (Section~\ref{sec:vistel}) and its associated knowledge $K_I$ is fetched. Then, we frame an instruction prompt with the question $Q$ and the knowledge $K_I$, and encode it using the ${\sc lmm}_{embedding}$ module $f$ to obtain textual features $X_{T_{Q:K_I}}$. We encode the image $I$ using a vision encoder to obtain visual features $X_I$. Then, we concatenate $X_I$ and $X_{T_{Q:K_I}}$ and feed them to the {\sc lmm} to generate an accurate answer $A$ to the question $Q$. Instruction prompt templates used in our ablation study are shown in the bottom right box, where $T_I^{ocr}$ is the visual text of the image $I$.}
  \label{fig:main_arch}
\end{figure*}

\noindent\textbf{Problem Statement:}  \data{}~\cite{singh2019strings} is a knowledge-intensive visual question-answering task that requires a system to read and interpret the visual text in an image and leverage it as a gateway to access and reason over external knowledge to answer the question. The external knowledge base $\mathcal{K}$ consists of a set of $n$ entities $\mathcal{E} = \{E_1, E_2, ..., E_n\}$ and their corresponding knowledge $\mathcal{K} = \{K_1, K_2, ..., K_n\}$, where each $K_i$ is a set of facts. For example, \emph{Domino's Pizza} is an entity whose associated knowledge facts, obtained in the form of triplets from Wikidata, are concatenated to form simple sentences such as \emph{“Domino's Pizza is a restaurant”, “Its headquarters are in Ann Arbor Charter Township”, “It belongs to the fast food industry”, and so on}. In this section, we describe our approach, whose overall architecture is illustrated in Figure~\ref{fig:main_arch}. Our approach first links visual text entities using the proposed \visTEL{} module and retrieves relevant knowledge to the entity (Section~\ref{sec:vistel}), it then reasons over the image and the retrieved knowledge to answer the question (Section~\ref{sec:kalma}). 
\begin{figure}[!t]
\centering
  \includegraphics[width=\columnwidth]{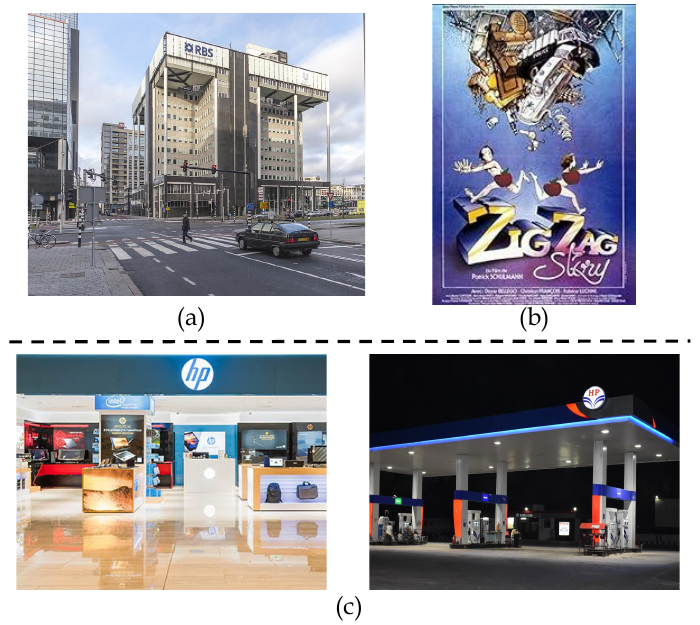}
  \caption{\textbf{Challenges associated with Visual Text Entity Linking:} (a) Visual text entity may appear as abbreviation instead of the entity name directly, e.g. ``RBS" instead of ``The Royal Bank of Scotland", (b) Visual text with varying font and stylized orientation pose a challenge to the recognizer, (c) Example of homonyms where visual text \emph{HP} may refer to  `Hewlett Packard' (left) or `Hindustan Petroleum' (right).}
  \label{fig:edit_dist_fig}
\end{figure}
\subsection{\visTEL{}: \underline{Vis}ual \underline{T}ext \underline{E}ntity \underline{L}inker}
\label{sec:vistel}
Entity linking is a well-studied task~\cite{10.5555/1214993}, where given a sentence, the named entities need to be identified and linked with their corresponding entities in a knowledge base. In this work, we study an analogous task, where the input is no longer a sentence, but instead an image containing visual text entities and the task is to link them to a corresponding external knowledge base. 

\begin{figure}[!t]
\centering
  \includegraphics[width=\columnwidth]{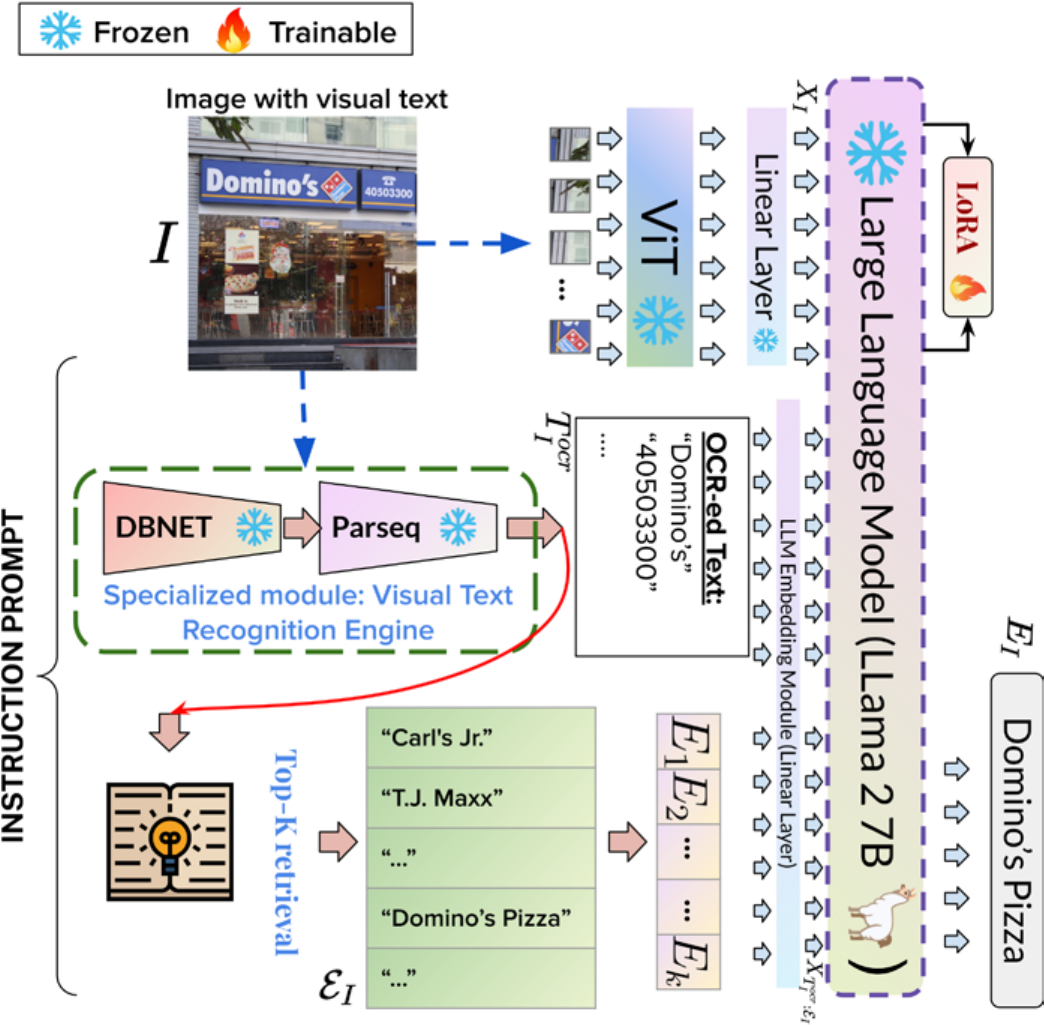}
  \caption{\textbf{Illustration of \visTEL{}.} We extract visual text from the given image using visual text recognition engine and, based on textual similarity, obtain $k$ candidate entities from the knowledge base. We fit OCRed text and the candidate entities into an instruction prompt template and encode the image using a visual encoder and the text prompt using an {\sc lmm} embedding module to obtain $X_I$ and $X_T$, respectively. Once encoded, {\sc lmm} generates the entity associated with the visual text in the image. Please refer to the Section~\ref{sec:vistel}.}
  \label{fig:vistel}
\end{figure}

One plausible solution, as shown in~\cite{singh2019strings}, is to extract the visual text in these images using visual text recognition engines and then leverage distance-based text similarity methods between the recognized text and the candidate entities for the entity linking task. However, such methods are highly sensitive to the following challenges: (i) Noisy or imperfect OCR may lead to wrong entity linking, and (ii) visual text might contain abbreviations instead of the entity names, e.g. \emph{``RBS'}' for the entity \emph{``The Royal Bank of Scotland''}, (iii) The problem of homonymy, e.g. visual text \emph{HP} may refer to \emph{`Hindustan Petroleum'} or \emph{`Hewlett Packard'}. Furthermore, unlike entity linking which often benefits from larger textual contexts; visual text entity linking has limited textual context, e.g., surrounding visual texts, and often must infer correct entities based on visual context. Please refer to Figure~\ref{fig:edit_dist_fig} for a selection of challenges associated with visual text entity linking. The other plausible solution is to use large multimodal models ({\sc lmm}s). By virtue of large-scale pretraining, they have strong abilities to reason and infer correct entities based on visual cues. However, we observe that feeding only the image without the surrounding OCRed text often results in hallucinations. To address these shortcomings, we propose \underline{Vis}ual \underline{T}ext \underline{E}ntity \underline{L}inker (\visTEL) that links the visual text present in an input image to its corresponding entity by jointly reasoning on textual context obtained using an explicit specialized visual text recognition engine and visual context obtained using a vision encoder of a large multimodal model. The architecture for \visTEL{} is illustrated in Figure~\ref{fig:vistel}. 

\noindent \textbf{Visual Text Recognition Engine:} Given an image $I$, we extract text $T^{ocr}_I = \{t^{ocr}_1, t^{ocr}_2, ..., t^{ocr}_r\}_I$ using specialized visual text detection and recognition methods. We, then find a set of $k$ candidate entities $\mathcal{E}_I$ based on the normalized edit-distance (NED) score between the entity name in the knowledge base with $T_I^{ocr}$. We use state-of-the-art text detection and text recognition approaches, namely {\sc dbnet}~\cite{dbnet} and ParSeq~\cite{parseq}, respectively.

\noindent \textbf{Vision encoder:} We use the output of the last transformer layer of a pretrained CLIP visual encoder ViT-L/14~\cite{radford2021learning} as our patched image features $\Tilde{X}_I \in \mathbb{R}^{p\times d_{v}}$, where $p$ and $d_v$ are the number of patches and encoding dimension of ViT, respectively. Further, these image features are projected to $d_{lmm}$ dimension using a linear layer $g$ to obtain the final sequence of image features $X_I \in \mathbb{R}^{p \times d_{{\sc lmm}}}$, i.e., $X_I = g(\Tilde{X}_{I})$.

\noindent\textbf{Large Multimodal Model:} Once we obtain the OCR-ed text $T_I^{ocr}$ and candidate entities $\mathcal{E}_I$, we frame the following instruction prompt:
\begin{tcolorbox}[title=Instruction prompt template for \visTEL{}] 
\footnotesize
\textcolor{blue}{$<$image$>$}\\
USER:{Given an image. The task is to link the visual text \textcolor{blue}{\{$T_I^{ocr}$\}} to one of the following entities: \textcolor{blue}{\{$\mathcal{E}_I$\}}}\\
ASSISTANT:\textcolor{red}{\{$E_I$\}}
\label{prompt_vistel}
\end{tcolorbox}

Then, we feed the prompt to the embedding module $h$ of the {\sc lmm} to obtain text tokens $X_{T^{ocr}_I:\mathcal{E}_I} \in \mathbb{R}^{l \times d_{{\sc lmm}}}$ i.e., $X_{T^{ocr}_I:\mathcal{E}_I} = h(prompt(T^{ocr}_I:\mathcal{E}_I))$, where $l$ and  $d_{lmm}$ are the number of text tokens and input embedding dimension for the {\sc lmm}, respectively. We, then concatenate image features $X_I$ and text features $X_{T^{ocr}_I:\mathcal{E}_I}$, and feed it as an input to the large multimodal model. \visTEL{} auto-regressively predicts the probability of the next token $E_{I_t}$ in the target entity $E_I$ by attending to the input prompt tokens and the previously generated entity tokens $E_{I<t}$. We train \visTEL{} by optimizing the language modeling loss for generating the target entity conditioned on the inputs $X_I$ and $X_{T^{ocr}_I:\mathcal{E}_I}$. 

\subsection{KaLMA: \underline{K}nowledge-\underline{a}ware \underline{L}arge \underline{M}ultimodal \underline{A}ssistant}
\label{sec:kalma}
We present Knowledge-aware Large Multimodal Assistant (\model{}) for addressing \data{}. The \model{} is an effective architecture that seamlessly integrates questions and images in the context of external knowledge in a trainable architecture to generate accurate answers.

We use visual features $X_I$ from the vision encoder. Further, we concatenate question $Q$ and the knowledge $K_I$ via instruction prompt template (as shown in the Figure~\ref{fig:main_arch}) and feed to the embedding module $f$ of the {\sc lmm} to obtain text tokens $X_{T_{Q:K_I}} \in \mathbb{R}^{m \times d_{lmm}}$ i.e., $X_{T_{Q:K_I}} = f(prompt(Q:K_I))$, where m is the number of text tokens. Then, we concatenate image features $X_I$, and text features $X_{T_{Q:K_I}}$ and feed to the large multimodal model to generate the accurate answer $A$. Further, to bring attribution ability, we model \model{} to generate the supporting fact $S$ that contributed to the answer along with answer generation. From here onwards, we will refer answer and supporting fact together as $A$. \model{} predicts the probability of the next token $A_{a_t}$ in the answer $A_a$ in an auto-regressive manner. It does so by attending to the prompt inputs and the previously generated tokens $A_{a<t}$. We train by minimizing the generative language modeling loss $\mathcal{L}_{ans\_gen}(\theta)$, which aims to generate the target tokens based on the inputs $X_I$ and $X_{T_{Q:K_I}}$ (Eq.~\ref{eqn:loss}). Note that target tokens comprise both the answer and the supporting fact. During training, we leverage the ground truth entity and its corresponding knowledge $K_I$, while during inference, we obtain it using our \visTEL{} module. We reuse the weights of \visTEL{} to initialise \model{}.

\begin{equation}
    \label{eqn:loss}
    \scriptsize
    \mathcal{L}_{ans\_gen}(\theta) =  -\left[\sum_{t=1}^{|A|}\log(P_\theta(A_{a_t}|A_{a<t}, X_I, X_{T_{Q:K_I}}))\right],
\end{equation}

where $\theta$ are the trainable parameters, $A_{a<t}$ represents the answer tokens already generated before predicting the token $A_{a_t}$ at the current time step $t$. 

\section{Experiments and Results}

\begin{table}[t]
  \centering
 \resizebox{1\columnwidth}{!}
 {
  \begin{tabular}{l c c c }
  \toprule
  \multicolumn{1}{c}{} & \multicolumn{3}{c}{Accuracy on \data{}} \\
  \cmidrule(r){2-4}
  \multicolumn{1}{l}{Method} & \multicolumn{1}{c}{scene} & \multicolumn{1}{c}{book} & \multicolumn{1}{c}{movie} \\

  \midrule
  \textbf{Traditional VQA Baselines} & & & \\
   ~BiLSTM & 17.0 & 12.4 & 11.3 \\
   ~BoW$+$CNN & 11.5 & 8.7 & 7.0 \\
   ~BLSTM$+$CNN~\cite{antol2015vqa}  & 19.8 & 17.3 & 15.7 \\
   ~HiCoAttenVQA~\cite{lu2016hierarchical} & 22.2 & 20.4 & 18.4\\
   ~BAN~\cite{DBLP:conf/nips/KimJZ18} & 23.5 & 22.3 & 20.3 \\
     \midrule
   \textbf{Pre-LLM Approaches}\\
   ~GPT-2~\cite{gpt2} & 22.8 & 22.3 & 31.8\\
   ~GPT-2 (w/ Visual Context) & 25.4 & 43.2 & 38.5 \\
   ~ViLT~\cite{vilt} & 38.2 & 31.1 & 40.1 \\
   ~VLBart~\cite{cho2021vlt5} & 35.1 & 38.6 & 41.5 \\
   \midrule
   \textbf{Previous SOTA} & & & \\
    ~{Memory Network~\cite{weston2015memory}}  & 49.0 & 57.2 & 42.0 \\
    ~{Singh et al.~\cite{singh2019strings}} & 54.5 & 62.7 & 45.2 \\
   \midrule
   \textbf{LLM-based Approaches}\\
   ~mPlug-Owl~\cite{ye2023mplug} & 21.3 & 26.7 & 8.2 \\
   ~LLaVA-1.5~\cite{liu2023llava} & 39.2 & 37.0 & 46.1 \\
   ~MiniGPT4v2~\cite{zhu2023minigpt} & 48.2 & 47.7 & 47.6 \\
   ~InstructBLIP~\cite{dai2024instructblip} & 31.5 & 30.3 & 29.9 \\
    \midrule
   \textbf{Ours (\model{})} &  &  & \\
    ~{w/ NED retrieval} & 54.9 & 63.4 & 70.8 \\
    ~{w/ \visTEL{}} & \textbf{72.7} ($\uparrow$ 18.2\%) & \textbf{82.3} ($\uparrow$ 19.6\%) & \textbf{77.4} ($\uparrow$ 32.2\%) \\
    \cdashline{1-4} 
    ~{Oracle} & 99.3 & 92.8 & 99.4  \\

  \bottomrule 
\end{tabular}
}

\caption{\textbf{Results on \data{}:} Various methods on the three data categories of \data{} dataset, namely, scene, book and movie.}
\label{tab:main_result}
\end{table}

\subsection{Dataset, Metrics and Comparisons}
We conduct our experiments on \data{}~\cite{singh2019strings} dataset\footnote{Available at: \url{https://textkvqa.github.io}}. The questions in this dataset span across three splits, namely, scene,  book, and movie containing natural scene images, book covers, and movie posters, respectively. These splits have (50K questions, 10K images, 500 entities), (1M questions, 207K images, 207K entities), (222K questions, 34K images, 34K entities), respectively. Further, each of these splits comes with its own knowledge base, namely \emph{KB-business} containing knowledge facts about business brand entities harvested from Wikidata, \emph{KB-book} containing knowledge facts about books harvested from a book catalog, and \emph{KB-movie} containing knowledge facts about movies harvested from IMDB, respectively. For each split, we follow the similar train-test division as~\cite{singh2019strings} where entities in train and test sets are disjoint. We evaluate the methods using an accuracy metric. 

\begin{figure*}[t!]
\centering
\begin{subfigure}[b]{\textwidth}
   \includegraphics[width=1\textwidth]{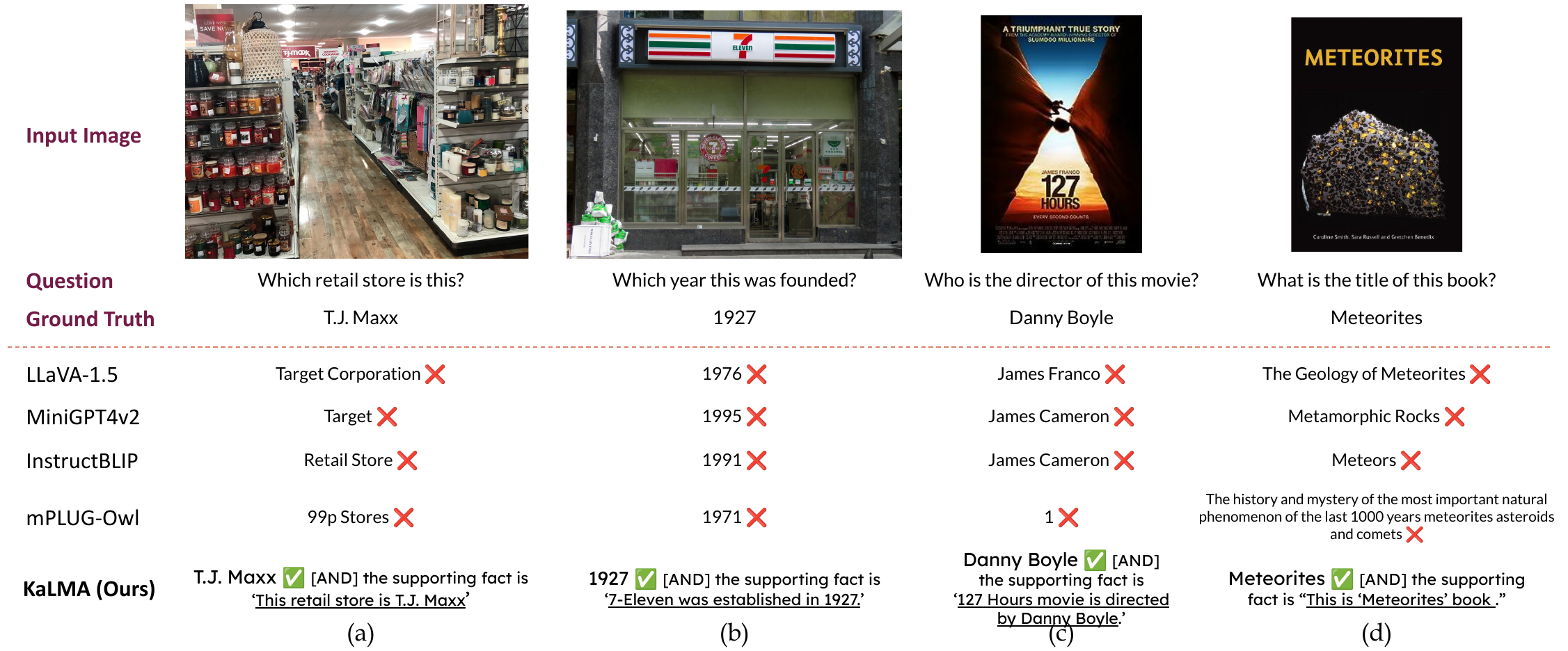}
\end{subfigure}
\begin{subfigure}[b]{\textwidth}
   \includegraphics[width=1\textwidth]{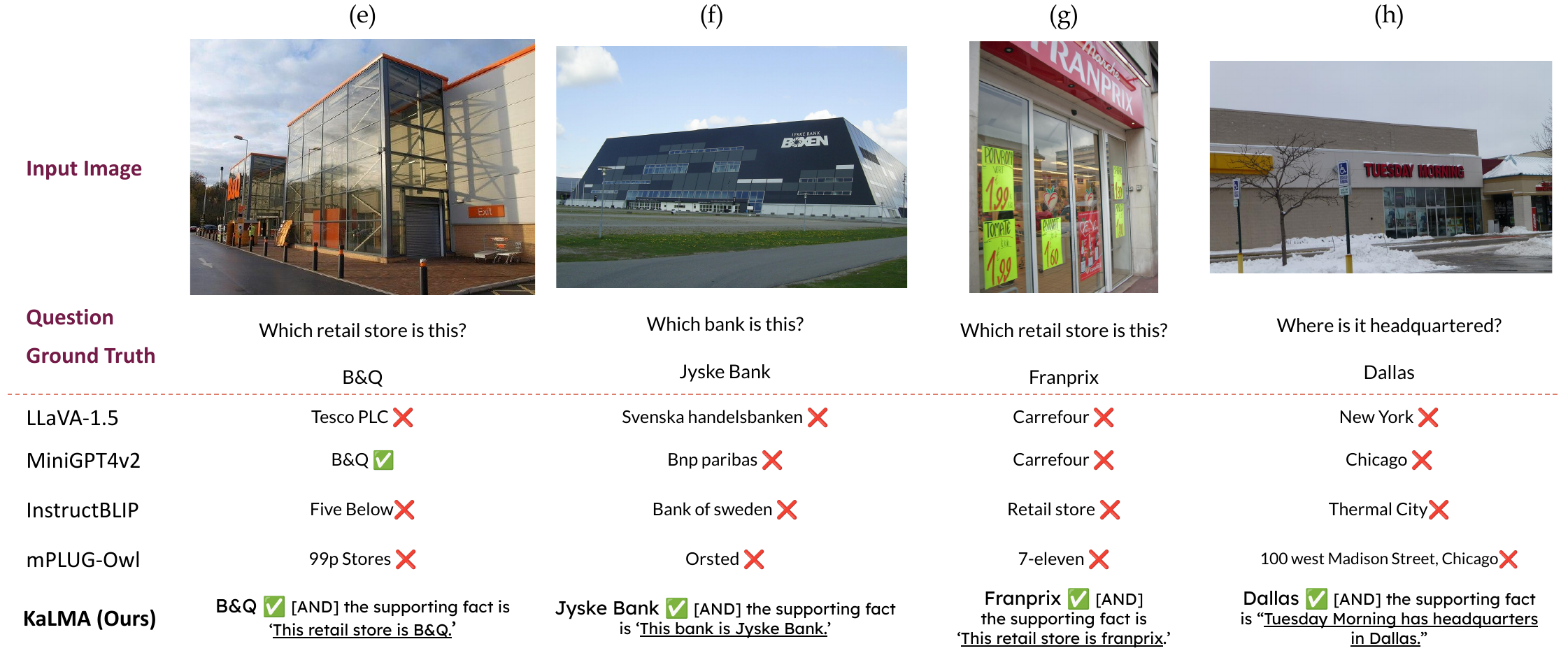}
\end{subfigure}
\caption{\textbf{A selection of our results as compared to implicit knowledge-based {LMM} approaches.} Please refer Qualitative Results in Section~\ref{sec:results} for observations. More results in Appendix~\ref{sec:app_more_results}.}
\label{fig:select_results}
\end{figure*}

Along with traditional VQA baselines, we compare the question answering performance of our proposed approach \model{} with methods from the following three major categories: (i) \textbf{Pre-{LMM} Approaches}: here, we choose classical transformer-based baselines, namely, GPT-2~\cite{gpt2} (text-only), GPT-2 (with BLIP-2~\cite{blip2}-extracted captions as visual context), ViLT~\cite{vilt} and VLBart~\cite{cho2021vlt5}. For an encoder-only model like ViLT, we treat \data{} as a classification-style visual question answering where the task is to predict the answer from a set of all possible answers.
(ii) \textbf{{LMM}-based Approaches:} restricting ourselves to open-source models, we choose four popular {\sc lmm}s, namely, mPlug-Owl~\cite{ye2023mplug}, MiniGPT4v2~\cite{zhu2023minigpt}, LLaVA-1.5 (7B)~\cite{liu2023llava} and InstructBLIP~\cite{dai2024instructblip} for comparison. Prompts used and other fine-tuning details for these {\sc lmm}s are discussed in the Appendix. (iii) \textbf{SOTA approaches:} we also compare against memory network~\cite{weston2015memory} and graph neural network-based approach~\cite{singh2019towards} which are the current state of the art. In addition to these comparisons, we compare the visual text entity linking performance of our proposed \visTEL{} against recent multimodal retrievers from UniIR~\cite{wei2023uniir}, specifically CLIP-SF and BLIP-SF, where we use image and visual text to retrieve entities from the knowledge base.

\subsection{Implementation Details}
We implemented our method using PyTorch and the Huggingface Transformers library~\cite{wolf-etal-2020-transformers}. We used LLaVA-1.5 as our foundation model for both \visTEL{} and \model{} models. Note that, LLaVA-1.5 is trained on CC3M~\cite{cc3m} and MS-COCO~\cite{coco}. We have carefully examined these datasets for duplicates and found no overlap with the evaluation set of \data{}. Further, {\sc dbnet}~\cite{dbnet} and {\sc ParSeq}~\cite{parseq} are used as visual-text detection and visual-text recognition modules in the visual text recognition engine, respectively. We fine-tuned \visTEL{} with LoRA for 10 epochs with a learning rate of 1e-5 with a batch size of 128. Similarly, we fine-tuned \model{} with LoRA for 6 epochs with a learning rate of 2e-5 with a batch size of 64. LoRA details are as follows: rank: 16, alpha: 32, dropout: 0.05, for both the models. Our experiments are conducted on a machine with three A6000 GPUs (48 GB each). We make our implementation publicly available at our project website\footnote{\scriptsize \url{https://vl2g.github.io/projects/LMM4Text-KVQA/}}. 

\subsection{Results and Discussion}
\label{sec:results}
\textbf{Results on \data{}:} We quantitatively evaluate our proposed framework \model{} on \data{} and compare against relevant methods in Table~\ref{tab:main_result}. We report accuracy averaged over the entire test set for all the three splits of \data{}. It is no surprise that traditional VQA baselines perform poorly as they do not have the ability to read and reason over visual text. Pre-{\sc lmm} language models (GPT-2) and vision-language models (GPT-2 w/ visual context, ViLT, VisualBert) along with {\sc lmm} baselines (mPlug-Owl, LLaVA-1.5, MiniGPT4v2, InstructBLIP) outperform traditional methods, but fail to outperform knowledge-aware methods including the state-of-the-art method~\cite{singh2019towards}. We observe that on knowledge-intensive tasks like \data{}, the OCR-free capabilities acquired by {\sc lmm}s are due to heavily correlated hallucinations of visual objects, thereby fall short to our proposed approach by a significant margin. Our proposed framework seamlessly integrates knowledge associated with visual text entity (extracted using our proposed \visTEL{}) and significantly enhances the performance on \data{}. To be specific, we advance the state-of-the-art by 18.2\%, 19.6\%, and 32.2\% on scene, book, and movie splits of \data{} on an absolute scale. This superiority of our approach demonstrates its efficacy in knowledge-aware text-based visual question answering. 

\begin{figure}[t!]
\centering
  \includegraphics[width=\columnwidth]{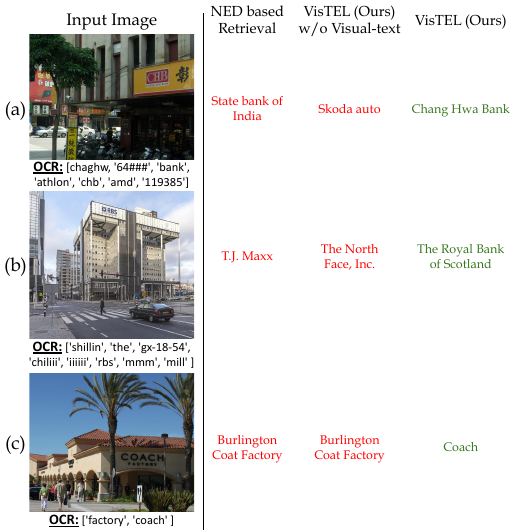}
  \caption{\textbf{Comparison of visual text entity linking results.} The \visTEL{} infers the correct entity based on visual context as well as textual context in the form of surrounding text in the image. Please refer to Qualitative Results in Section~\ref{sec:results} for more details.}
  \label{fig:vel_exps}
\end{figure}
\noindent\textbf{Visual Text Entity Linking Results:} We report them in Table~\ref{tab:vel_results}. Here, we observe that the proposed \visTEL{} clearly outperforms both (i) Text-only retrievers, such as a direct match or normalized edit distance-based match of OCRed text and entity name, and (ii) Multimodal retrievers, CLIP-SF and BLIP-SF from UniIR~\cite{wei2023uniir}. By virtue of {\sc lmm} and joint reasoning of visual and textual (OCR) context for linking visual text, \visTEL{} yields reasonably advanced performance. Nevertheless, there is still scope of improvement which we believe can be achieved by further improving visual text recognition, and performing detailed visual reasoning such as logo recognition. We leave these extensions as future work.

\begin{table}[!t]
\centering
    \scriptsize
    \resizebox{\columnwidth}{!}
    {
    \begin{tabular}{lccccc}
    \toprule
    Method & Visual Context & Textual Context & scene & book & movie\\
    \midrule
   \textbf{Text-only} & & & & & \\
    ~~~Direct match & \xmark & \cmark & 54.8 & 63.6 & 58.1\\
    ~~~NED & \xmark & \cmark & 57.1 & 66.5 & 60.1\\
    \midrule
    \textbf{Multimodal retrievers} & & & & & \\
    ~~~UniIR (\textit{CLIP-SF}) & \cmark & \cmark & 64.5 & 78.8 & 45.2 \\
    ~~~UniIR (\textit{BLIP-SF}) & \cmark & \cmark & 60.6 & 78.5 & 50.1 \\
    \cdashline{1-6}
    ~~~\textbf{Ours} & & & & & \\
    ~~~~\visTEL{} & \cmark & \xmark & 73.2 & 76.9 & 66.6 \\
    ~~~~\visTEL{} & \xmark & \cmark & 31.5 & 9.8 & 11.6\\
    ~~~~\visTEL{} & \cmark & \cmark & \textbf{76.5} & \textbf{80.6} & \textbf{71.6} \\

    \bottomrule
  \end{tabular}}
    \caption{\textbf{Visual Text Entity Linking Results.} We report Recall@1. Text-only retrievers: direct match and normalized edit distance-based methods and Multimodal retrievers: CLIP-SF and BLIP-SF from UniIR~\cite{wei2023uniir} fall short. On the contrary, the proposed \visTEL{}, which leverages both visual and textual context (surrounding OCRed text) in an {\sc lmm} framework, shows impressive visual text entity linking performance over both text-only as well as multimodal retrievers.}
  \label{tab:vel_results}
\end{table}

\noindent\textbf{Qualitative Results:} We show a selection of results for text-based knowledge-aware visual question answering and visual text entity linking in Figure~\ref{fig:select_results} and Figure~\ref{fig:vel_exps}, respectively. 

In Figure~\ref{fig:select_results}, {\sc lmm} models exhibit hallucination over visually apparent objects. In (a), all {\sc lmm}s incorrectly identify \emph{T.J. Maxx} as popular retail stores \emph{Target} and \emph{99p Stores}. In (b), they provide a random year. In (c), these models are confused over the keyword \emph{James}, mixing up the director and actor names on the poster. In (d), {\sc lmm}s hallucinate and suggest non-existent book titles. Similar hallucinations can be seen in the other examples (e-h). Our proposed method owing to visual-text entity linking capabilities and reasoning over explicit knowledge, provides accurate answers.

In Figure~\ref{fig:vel_exps}, we observe that our proposed model accurately links visual text in the images to the correct entity despite noisy OCR in (a), abbreviations in (b), and ambiguous visual text in (c). 

\begin{table}[!t]
\centering
    \scriptsize
    \resizebox{\columnwidth}{!}
    {
    \begin{tabular}{c c c ccc}
    \hline
    Model & Visual text EL & Knowledge & scene & book & movie\\
    \midrule
    \multirow{5}{*}{\model{}} & \xmark & \xmark & 39.2 & 37.0 & 46.1 \\
    & \xmark & OCR only & 52.2 & 49.8 & 51.7 \\
    & \cmark & Entity name only & 53.2 & 59.1 & 59.2 \\
    & \cmark (w/o \visTEL{}) & Knowledge facts & 54.9 & 63.4 & 70.8 \\
    & \cmark (w/ \visTEL{}) & Knowledge facts & \textbf{72.7} & \textbf{82.3} & \textbf{77.4} \\
    \cdashline{1-6}
    \makecell{MiniGPT4v2 \\ (best LMM method)} & \xmark & \xmark & 48.2 & 47.7 & 47.6 \\
    \hline
  \end{tabular}}
    \caption{Ablations for showing the importance of visual text entity linking, explicit knowledge facts and \visTEL{}. Also, note that the first-row result corresponds to LLaVA-1.5 result from Table~\ref{tab:main_result}, as \model{} without \visTEL{} and knowledge is equivalent to LLaVA-1.5. Please refer to Section~\ref{sec:abl} for more details. }
  \label{tab:ent_ocr_exp}
\end{table}

\subsubsection{Ablations and Analysis}
\label{sec:abl}
We conduct the following ablations and analysis of the proposed work:

\noindent\textbf{\emph{(i) What is the need for \visTEL{}?:}} To study the performance of our model in the absence of the proposed \visTEL{} module, we replace it with traditional edit-distance-based entity linking where entities are sorted based on the normalized edit-distance between extracted OCRs and the entity name. The results of this ablation, as shown in Table~\ref{tab:ent_ocr_exp} further support our claim that the superior visual text entity linking capabilities of the proposed \visTEL{}, enhances the downstream performance of \model{}.

\noindent\textbf{\emph{(ii) What is the need for visual text entity linking and explicit knowledge in \data{}?:}}  We show these ablation results in Table ~\ref{tab:ent_ocr_exp}. We, \textbf{first}, skip visual entity linking in the \model{}, and feed only the extracted OCRed text to \model{}. The drop in performance shows the utility of visual text entity linking. \textbf{Second}, we perform visual entity linking, but, we feed the visual text linked entity name from the \visTEL{} as input to \model{}. Our observations indicate that although entity names give some hints about the associated knowledge and reduce hallucination to some extent, it is not as useful as using explicit knowledge in our full model. 

\noindent\textbf{\emph{(iii) How much does choice of visual text recognition engine matter?:}} In this ablation, we replace {\sc dbnet}~\cite{dbnet} and  {\sc p}ar{\sc s}eq~\cite{parseq} used in \model{} with {\sc craft}~\cite{craft}, {\sc east}~\cite{zhou2017east} and {\sc crnn}~\cite{crnn}, and report the results of \model{} on \data{} in Table~\ref{tab:qa_diff_det_rec}. Although effective visual text recognition is critical to the performance, our model that jointly reasons on visual and textual context, performs reasonably well even with sub-par visual text recognition. 

\begin{table}[!t]
\centering
    \scriptsize
    \resizebox{0.8\columnwidth}{!}
    {
    \begin{tabular}{ll  ccc}
    \hline
    Detection & Recognition & scene & book & movie\\
    \midrule
    EAST & CRNN & 67.2 & 81.3 & 66.4\\
    CRAFT & CRNN & 67.7 & 81.9 & 75.1 \\
    DBNet & ParSeq & \textbf{72.7} & \textbf{82.3} & \textbf{77.4} \\      
    \hline
  \end{tabular}}
    \caption{Effect of Different Text Detection and Recognition Approaches in our approach.}
  \label{tab:qa_diff_det_rec}
\end{table}

\noindent\textbf{\emph{(iv) Attribution ability of \model{}}:}
To study the impact of support fact generation (SFG) along with the answer generation on the performance of \model{}, we train \model{} without support fact generation, and report the results in Table~\ref{tab:wo_sfg}. We observe that \model{}'s performance drops slightly, further supporting our claim that support fact generation elicits chain-of-thought reasoning, thereby improving the performance of answer generation along with adding attribution abilities to the model.

\noindent\textbf{\emph{(iv) Cost analysis:}}
We provide a comparison of \model{} with a traditional non-LLM-based approach (ViLT). Our approach takes on average 5.6s per sample, which includes 4s for visual text recognition, 0.8s for entity linking using VisTEL and 0.8s for VQA using KaLMA as compared to ViLT which takes on average 0.2s per sample during inference. The training time (finetuning) of both these models are 36 and 8 hrs, respectively. Furthermore, the trainable parameters for both these models are 20M (Total size: 14B) and 114M (Total size: 114M), respectively. We achieved speed-up in our LMM components through parameter-efficient fine-tuning (LoRA) with 16-bit precision and 8-bit quantization during inference. As anticipated, traditional models have a notable advantage in terms of computational efficiency compared to our LMM-based approach. Nonetheless, we substantially surpass them in \data{} accuracy.

\section{Conclusion}
\label{sec:conc}
\begin{table}[!t]
\centering
    \scriptsize
    \resizebox{0.6\columnwidth}{!}
    {
   \begin{tabular}{l c ccc}
   \hline
   SFG &scene & book & movie\\
   \midrule
   \cmark & \textbf{72.7} & 82.3 & \textbf{77.4} \\
   \xmark & 71.4 & \textbf{83.5} & 76.9 \\
   \hline
 \end{tabular}}
   \caption{Performance of \model{} w/ and w/o supporting fact generation (SFG) on \data{}.}
  \label{tab:wo_sfg}
\end{table}
We have revisited the \data{} and significantly advanced state of the art on this task. Our findings suggest that visual text entity linking, combined with seamless reasoning using both visual and textual cues, as well as explicit external knowledge via {\sc lmm}, is key to our success. We performed extensive ablation studies and analyses to support our claims.  The future scope of this work is to expand the dataset with more visual-intensive queries and address \data{} for multilingual societies. 

\section{Limitations}
We observe the following limitations in our work: (i) Existing visual text recognition pipelines suffer on low-resolution images where it is challenging to extract visual text, which further impacts the performance of our \visTEL{} (ii) In the dataset we use, it was assumed that each image contains only one visual text entity which may not be always true in a real-world scenario. (iii) Current state-of-the-art visual text recognition engines are not effective enough over multi-lingual text in the wild; Hence, in this work, we further assume the visual-text is English which again might not hold in a realistic setting. (iv) The temporal nature of knowledge, such as the entity ``Statoil" being renamed ``Equinor" over time, is not handled by our current models. We leave addressing these limitations as a future work of this paper. 

\section{Ethical Considerations and Broader Impact}
This work is based on the publicly available \data{} dataset, which predominantly contains English visual text, and the associated knowledge base, questions, and answer pairs are also in English. The dataset may have some geographic bias that went undetected in this work, a common issue with many public computer vision and NLP benchmarks. Additionally, our work uses large multimodal models ({\sc lmm}s), which can inherit and potentially amplify biases from the large-scale pretraining data used.

We are mindful of the environmental impact of using {\sc lmm}s due to their heavy computational requirements. To mitigate this, we judiciously used {\sc lmm}s by reusing pre-existing checkpoints wherever appropriate.

We open-source our implementation to facilitate reproduction and further study. Nevertheless, a more rigorous inspection is indeed required before deploying the proposed model in real-world applications to ensure ethical considerations are comprehensively addressed.

\noindent\textbf{Broader Impact}:
The proposed work has the following broader impact: (i) The ability to link visual text entities to knowledge bases and leverage this linked knowledge for answering questions can improve the accuracy and relevance of information retrieval systems. Although not studied in this work, this may be particularly valuable in content recommendation systems and search engines. (ii) This research contributes to advancing the capabilities of AI systems to understand and interact with multimodal information (text and images), which can benefit applications in fields such as virtual assistants, content understanding, and automated decision-making. (iii) Methodologically, contributions such as \visTEL{} provide new frameworks and techniques for visual text entity linking, which can inspire further innovations in Visual NLP. 

\section*{Acknowledgements}
This work was partly supported by the IIT Jodhpur Seed Research Grant and National Language Translation Mission (NLTM): Bhashini project by the MeitY, Government of India. Abhirama Subramanyam Penamakuri was supported by the PMRF fellowship, MoE, Government of India.

\bibliography{custom}

\begin{thebibliography}{62}
\providecommand{\natexlab}[1]{#1}

\bibitem[{Antol et~al.(2015)Antol, Agrawal, Lu, Mitchell, Batra, Zitnick, and Parikh}]{antol2015vqa}
Stanislaw Antol, Aishwarya Agrawal, Jiasen Lu, Margaret Mitchell, Dhruv Batra, C~Lawrence Zitnick, and Devi Parikh. 2015.
\newblock {VQA}: Visual question answering.
\newblock In \emph{ICCV}.

\bibitem[{Baek et~al.(2019)Baek, Lee, Han, Yun, and Lee}]{craft}
Youngmin Baek, Bado Lee, Dongyoon Han, Sangdoo Yun, and Hwalsuk Lee. 2019.
\newblock Character region awareness for text detection.
\newblock In \emph{CVPR}.

\bibitem[{Bautista and Atienza(2022)}]{parseq}
Darwin Bautista and Rowel Atienza. 2022.
\newblock Scene text recognition with permuted autoregressive sequence models.
\newblock In \emph{ECCV}. Springer.

\bibitem[{Biten et~al.(2019{\natexlab{a}})Biten, Tito, Mafla, Gomez, Rusinol, Mathew, Jawahar, Valveny, and Karatzas}]{biten2019icdar}
Ali~Furkan Biten, Ruben Tito, Andres Mafla, Lluis Gomez, Mar{\c{c}}al Rusinol, Minesh Mathew, CV~Jawahar, Ernest Valveny, and Dimosthenis Karatzas. 2019{\natexlab{a}}.
\newblock Icdar 2019 competition on scene text visual question answering.
\newblock In \emph{ICDAR}.

\bibitem[{Biten et~al.(2019{\natexlab{b}})Biten, Tito, Mafla, Gomez, Rusinol, Valveny, Jawahar, and Karatzas}]{biten2019scene}
Ali~Furkan Biten, Ruben Tito, Andres Mafla, Lluis Gomez, Mar{\c{c}}al Rusinol, Ernest Valveny, CV~Jawahar, and Dimosthenis Karatzas. 2019{\natexlab{b}}.
\newblock Scene text visual question answering.
\newblock In \emph{ICCV}.

\bibitem[{Brown et~al.(2020)Brown, Mann, Ryder, Subbiah, Kaplan, Dhariwal, Neelakantan, Shyam, Sastry, Askell et~al.}]{gpt3}
Tom Brown, Benjamin Mann, Nick Ryder, Melanie Subbiah, Jared~D Kaplan, Prafulla Dhariwal, Arvind Neelakantan, Pranav Shyam, Girish Sastry, Amanda Askell, et~al. 2020.
\newblock Language models are few-shot learners.
\newblock \emph{NeurIPS}.

\bibitem[{Caron et~al.(2024)Caron, Iscen, Fathi, and Schmid}]{caron2024generative}
Mathilde Caron, Ahmet Iscen, Alireza Fathi, and Cordelia Schmid. 2024.
\newblock A generative approach for wikipedia-scale visual entity recognition.
\newblock In \emph{CVPR}.

\bibitem[{Chen et~al.(2023)Chen, Hu, Luan, Sun, Changpinyo, Ritter, and Chang}]{infoseek}
Yang Chen, Hexiang Hu, Yi~Luan, Haitian Sun, Soravit Changpinyo, Alan Ritter, and Ming-Wei Chang. 2023.
\newblock Can pre-trained vision and language models answer visual information-seeking questions?
\newblock In \emph{EMNLP}.

\bibitem[{Cho et~al.(2021)Cho, Lei, Tan, and Bansal}]{cho2021vlt5}
Jaemin Cho, Jie Lei, Hao Tan, and Mohit Bansal. 2021.
\newblock Unifying vision-and-language tasks via text generation.
\newblock In \emph{ICML}.

\bibitem[{Chung et~al.(2022)Chung, Hou, Longpre, Zoph, Tay, Fedus, Li, Wang, Dehghani, Brahma et~al.}]{flant5}
Hyung~Won Chung, Le~Hou, Shayne Longpre, Barret Zoph, Yi~Tay, William Fedus, Yunxuan Li, Xuezhi Wang, Mostafa Dehghani, Siddhartha Brahma, et~al. 2022.
\newblock Scaling instruction-finetuned language models.
\newblock \emph{arXiv preprint arXiv:2210.11416}.

\bibitem[{Dai et~al.(2024)Dai, Li, Li, Tiong, Zhao, Wang, Li, Fung, and Hoi}]{dai2024instructblip}
Wenliang Dai, Junnan Li, Dongxu Li, Anthony Meng~Huat Tiong, Junqi Zhao, Weisheng Wang, Boyang Li, Pascale~N Fung, and Steven Hoi. 2024.
\newblock Instructblip: Towards general-purpose vision-language models with instruction tuning.
\newblock \emph{NeurIPS}.

\bibitem[{Devlin et~al.(2019)Devlin, Chang, Lee, and Toutanova}]{devlin-etal-2019-bert}
Jacob Devlin, Ming-Wei Chang, Kenton Lee, and Kristina Toutanova. 2019.
\newblock {BERT}: Pre-training of deep bidirectional transformers for language understanding.
\newblock In \emph{NAACL-HLT}.

\bibitem[{Gard{\`e}res et~al.(2020)Gard{\`e}res, Ziaeefard, Abeloos, and Lecue}]{garderes-etal-2020-conceptbert}
Fran{\c{c}}ois Gard{\`e}res, Maryam Ziaeefard, Baptiste Abeloos, and Freddy Lecue. 2020.
\newblock {C}oncept{B}ert: Concept-aware representation for visual question answering.
\newblock In \emph{EMNLP}.

\bibitem[{Goyal et~al.(2017)Goyal, Khot, Summers{-}Stay, Batra, and Parikh}]{goyal2017making}
Yash Goyal, Tejas Khot, Douglas Summers{-}Stay, Dhruv Batra, and Devi Parikh. 2017.
\newblock Making the {V} in {VQA} matter: Elevating the role of image understanding in visual question answering.
\newblock In \emph{CVPR}.

\bibitem[{Gui et~al.(2022)Gui, Wang, Huang, Hauptmann, Bisk, and Gao}]{gui2022kat}
Liangke Gui, Borui Wang, Qiuyuan Huang, Alexander~G Hauptmann, Yonatan Bisk, and Jianfeng Gao. 2022.
\newblock Kat: A knowledge augmented transformer for vision-and-language.
\newblock In \emph{NAACL-HLT}.

\bibitem[{Guu et~al.(2020)Guu, Lee, Tung, Pasupat, and Chang}]{realm}
Kelvin Guu, Kenton Lee, Zora Tung, Panupong Pasupat, and Mingwei Chang. 2020.
\newblock Retrieval augmented language model pre-training.
\newblock In \emph{ICML}.

\bibitem[{Hu et~al.(2023{\natexlab{a}})Hu, Luan, Chen, Khandelwal, Joshi, Lee, Toutanova, and Chang}]{HuLCKJLTC23}
Hexiang Hu, Yi~Luan, Yang Chen, Urvashi Khandelwal, Mandar Joshi, Kenton Lee, Kristina Toutanova, and Ming{-}Wei Chang. 2023{\natexlab{a}}.
\newblock Open-domain visual entity recognition: Towards recognizing millions of wikipedia entities.
\newblock In \emph{ICCV}.

\bibitem[{Hu et~al.(2023{\natexlab{b}})Hu, Iscen, Sun, Wang, Chang, Sun, Schmid, Ross, and Fathi}]{hu2023reveal}
Ziniu Hu, Ahmet Iscen, Chen Sun, Zirui Wang, Kai-Wei Chang, Yizhou Sun, Cordelia Schmid, David~A Ross, and Alireza Fathi. 2023{\natexlab{b}}.
\newblock Reveal: Retrieval-augmented visual-language pre-training with multi-source multimodal knowledge memory.
\newblock In \emph{CVPR}.

\bibitem[{Jurafsky and Martin(2009)}]{10.5555/1214993}
Daniel Jurafsky and James~H. Martin. 2009.
\newblock \emph{Speech and Language Processing (2nd Edition)}.
\newblock Prentice-Hall, Inc., USA.

\bibitem[{Khademi et~al.(2023)Khademi, Yang, Frujeri, and Zhu}]{mmreasoner}
Mahmoud Khademi, Ziyi Yang, Felipe~Vieira Frujeri, and Chenguang Zhu. 2023.
\newblock Mm-reasoner: A multi-modal knowledge-aware framework for knowledge-based visual question answering.
\newblock In \emph{EMNLP}.

\bibitem[{Kim et~al.(2018)Kim, Jun, and Zhang}]{DBLP:conf/nips/KimJZ18}
Jin{-}Hwa Kim, Jaehyun Jun, and Byoung{-}Tak Zhang. 2018.
\newblock Bilinear attention networks.
\newblock In \emph{NeurIPS}.

\bibitem[{Kim et~al.(2021)Kim, Son, and Kim}]{vilt}
Wonjae Kim, Bokyung Son, and Ildoo Kim. 2021.
\newblock Vilt: Vision-and-language transformer without convolution or region supervision.
\newblock In \emph{ICML}.

\bibitem[{Lewis et~al.(2020)Lewis, Perez, Piktus, Petroni, Karpukhin, Goyal, K{\"u}ttler, Lewis, Yih, Rockt{\"a}schel et~al.}]{rag}
Patrick Lewis, Ethan Perez, Aleksandra Piktus, Fabio Petroni, Vladimir Karpukhin, Naman Goyal, Heinrich K{\"u}ttler, Mike Lewis, Wen-tau Yih, Tim Rockt{\"a}schel, et~al. 2020.
\newblock Retrieval-augmented generation for knowledge-intensive nlp tasks.
\newblock \emph{NeurIPS}.

\bibitem[{Li et~al.(2023)Li, Li, Savarese, and Hoi}]{blip2}
Junnan Li, Dongxu Li, Silvio Savarese, and Steven Hoi. 2023.
\newblock Blip-2: bootstrapping language-image pre-training with frozen image encoders and large language models.
\newblock In \emph{ICML}.

\bibitem[{Liao et~al.(2020)Liao, Wan, Yao, Chen, and Bai}]{dbnet}
Minghui Liao, Zhaoyi Wan, Cong Yao, Kai Chen, and Xiang Bai. 2020.
\newblock Real-time scene text detection with differentiable binarization.
\newblock In \emph{AAAI}.

\bibitem[{Lin et~al.(2014)Lin, Maire, Belongie, Hays, Perona, Ramanan, Doll{\'a}r, and Zitnick}]{coco}
Tsung-Yi Lin, Michael Maire, Serge Belongie, James Hays, Pietro Perona, Deva Ramanan, Piotr Doll{\'a}r, and C~Lawrence Zitnick. 2014.
\newblock Microsoft coco: Common objects in context.
\newblock In \emph{ECCV}.

\bibitem[{Lin et~al.(2022)Lin, Xie, Chen, Xu, Zhu, and Yuan}]{lin2022revive}
Yuanze Lin, Yujia Xie, Dongdong Chen, Yichong Xu, Chenguang Zhu, and Lu~Yuan. 2022.
\newblock Revive: Regional visual representation matters in knowledge-based visual question answering.
\newblock \emph{NeurIPS}.

\bibitem[{Liu et~al.(2024)Liu, Li, Li, and Lee}]{liu2023llava}
Haotian Liu, Chunyuan Li, Yuheng Li, and Yong~Jae Lee. 2024.
\newblock Improved baselines with visual instruction tuning.
\newblock In \emph{CVPR}.

\bibitem[{Lu et~al.(2016)Lu, Yang, Batra, and Parikh}]{lu2016hierarchical}
Jiasen Lu, Jianwei Yang, Dhruv Batra, and Devi Parikh. 2016.
\newblock Hierarchical question-image co-attention for visual question answering.
\newblock In \emph{NeurIPS}.

\bibitem[{Marino et~al.(2021)Marino, Chen, Parikh, Gupta, and Rohrbach}]{marino2021krisp}
Kenneth Marino, Xinlei Chen, Devi Parikh, Abhinav Gupta, and Marcus Rohrbach. 2021.
\newblock Krisp: Integrating implicit and symbolic knowledge for open-domain knowledge-based {VQA}.
\newblock In \emph{CVPR}.

\bibitem[{Marino et~al.(2019)Marino, Rastegari, Farhadi, and Mottaghi}]{marino2019okvqa}
Kenneth Marino, Mohammad Rastegari, Ali Farhadi, and Roozbeh Mottaghi. 2019.
\newblock {OK-VQA}: A visual question answering benchmark requiring external knowledge.
\newblock In \emph{CVPR}.

\bibitem[{Mathew et~al.(2021)Mathew, Karatzas, and Jawahar}]{mathew2021docvqa}
Minesh Mathew, Dimosthenis Karatzas, and CV~Jawahar. 2021.
\newblock {D}oc{VQA}: A dataset for {VQA} on document images.
\newblock In \emph{WACV}.

\bibitem[{Mishra et~al.(2019)Mishra, Shekhar, Singh, and Chakraborty}]{mishra2019ocr}
Anand Mishra, Shashank Shekhar, Ajeet~Kumar Singh, and Anirban Chakraborty. 2019.
\newblock {OCR-VQA}: Visual question answering by reading text in images.
\newblock In \emph{ICDAR}.

\bibitem[{Narasimhan et~al.(2018)Narasimhan, Lazebnik, and Schwing}]{narasimhan2018out}
Medhini Narasimhan, Svetlana Lazebnik, and Alexander Schwing. 2018.
\newblock Out of the box: Reasoning with graph convolution nets for factual visual question answering.
\newblock \emph{NeurIPS}.

\bibitem[{Narasimhan and Schwing(2018)}]{narasimhan2018straight}
Medhini Narasimhan and Alexander~G Schwing. 2018.
\newblock Straight to the facts: Learning knowledge base retrieval for factual visual question answering.
\newblock In \emph{ECCV}.

\bibitem[{Ouyang et~al.(2022)Ouyang, Wu, Jiang, Almeida, Wainwright, Mishkin, Zhang, Agarwal, Slama, Ray et~al.}]{instructgpt}
Long Ouyang, Jeffrey Wu, Xu~Jiang, Diogo Almeida, Carroll Wainwright, Pamela Mishkin, Chong Zhang, Sandhini Agarwal, Katarina Slama, Alex Ray, et~al. 2022.
\newblock Training language models to follow instructions with human feedback.
\newblock \emph{NeurIPS}.

\bibitem[{Penedo et~al.(2024)Penedo, Malartic, Hesslow, Cojocaru, Alobeidli, Cappelli, Pannier, Almazrouei, and Launay}]{falcon}
Guilherme Penedo, Quentin Malartic, Daniel Hesslow, Ruxandra Cojocaru, Hamza Alobeidli, Alessandro Cappelli, Baptiste Pannier, Ebtesam Almazrouei, and Julien Launay. 2024.
\newblock The refinedweb dataset for falcon llm: Outperforming curated corpora with web data only.
\newblock \emph{NeurIPS}.

\bibitem[{Radford et~al.(2021)Radford, Kim, Hallacy, Ramesh, Goh, Agarwal, Sastry, Askell, Mishkin, Clark et~al.}]{radford2021learning}
Alec Radford, Jong~Wook Kim, Chris Hallacy, Aditya Ramesh, Gabriel Goh, Sandhini Agarwal, Girish Sastry, Amanda Askell, Pamela Mishkin, Jack Clark, et~al. 2021.
\newblock Learning transferable visual models from natural language supervision.
\newblock In \emph{ICML}.

\bibitem[{Radford et~al.(2019)Radford, Wu, Child, Luan, Amodei, Sutskever et~al.}]{gpt2}
Alec Radford, Jeffrey Wu, Rewon Child, David Luan, Dario Amodei, Ilya Sutskever, et~al. 2019.
\newblock Language models are unsupervised multitask learners.
\newblock \emph{OpenAI blog}, 1(8):9.

\bibitem[{Shah et~al.(2019)Shah, Mishra, Yadati, and Talukdar}]{shah2019kvqa}
Sanket Shah, Anand Mishra, Naganand Yadati, and Partha~Pratim Talukdar. 2019.
\newblock {KVQA}: Knowledge-aware visual question answering.
\newblock In \emph{AAAI}.

\bibitem[{Sharma et~al.(2018)Sharma, Ding, Goodman, and Soricut}]{cc3m}
Piyush Sharma, Nan Ding, Sebastian Goodman, and Radu Soricut. 2018.
\newblock Conceptual captions: A cleaned, hypernymed, image alt-text dataset for automatic image captioning.
\newblock In \emph{ACL}.

\bibitem[{Shi et~al.(2016)Shi, Bai, and Yao}]{crnn}
Baoguang Shi, Xiang Bai, and Cong Yao. 2016.
\newblock An end-to-end trainable neural network for image-based sequence recognition and its application to scene text recognition.
\newblock \emph{IEEE TPAMI}, 39(11):2298--2304.

\bibitem[{Singh et~al.(2019{\natexlab{a}})Singh, Mishra, Shekhar, and Chakraborty}]{singh2019strings}
Ajeet~Kumar Singh, Anand Mishra, Shashank Shekhar, and Anirban Chakraborty. 2019{\natexlab{a}}.
\newblock From strings to things: Knowledge-enabled {VQA} model that can read and reason.
\newblock In \emph{ICCV}.

\bibitem[{Singh et~al.(2019{\natexlab{b}})Singh, Natarajan, Shah, Jiang, Chen, Batra, Parikh, and Rohrbach}]{singh2019towards}
Amanpreet Singh, Vivek Natarajan, Meet Shah, Yu~Jiang, Xinlei Chen, Dhruv Batra, Devi Parikh, and Marcus Rohrbach. 2019{\natexlab{b}}.
\newblock Towards {VQA} models that can read.
\newblock In \emph{CVPR}.

\bibitem[{Sun et~al.(2022)Sun, Fan, Guo, Zhang, and Cheng}]{SunFG0C22}
Wen Sun, Yixing Fan, Jiafeng Guo, Ruqing Zhang, and Xueqi Cheng. 2022.
\newblock Visual named entity linking: {A} new dataset and {A} baseline.
\newblock In \emph{EMNLP (Findings)}.

\bibitem[{Touvron et~al.(2023)Touvron, Martin, Stone, Albert, Almahairi, Babaei, Bashlykov, Batra, Bhargava, Bhosale et~al.}]{touvron2023llama}
Hugo Touvron, Louis Martin, Kevin Stone, Peter Albert, Amjad Almahairi, Yasmine Babaei, Nikolay Bashlykov, Soumya Batra, Prajjwal Bhargava, Shruti Bhosale, et~al. 2023.
\newblock Llama 2: Open foundation and fine-tuned chat models.
\newblock \emph{arXiv preprint arXiv:2307.09288}.

\bibitem[{Tsimpoukelli et~al.(2021)Tsimpoukelli, Menick, Cabi, Eslami, Vinyals, and Hill}]{frozen}
Maria Tsimpoukelli, Jacob~L Menick, Serkan Cabi, SM~Eslami, Oriol Vinyals, and Felix Hill. 2021.
\newblock Multimodal few-shot learning with frozen language models.
\newblock \emph{NeurIPS}.

\bibitem[{Vaswani et~al.(2017)Vaswani, Shazeer, Parmar, Uszkoreit, Jones, Gomez, Kaiser, and Polosukhin}]{vaswani2017attention}
Ashish Vaswani, Noam Shazeer, Niki Parmar, Jakob Uszkoreit, Llion Jones, Aidan~N Gomez, {\L}ukasz Kaiser, and Illia Polosukhin. 2017.
\newblock Attention is all you need.
\newblock \emph{NeurIPS}.

\bibitem[{Wang et~al.(2017{\natexlab{a}})Wang, Wu, Shen, Dick, and Van Den~Hengel}]{wang2017fvqa}
Peng Wang, Qi~Wu, Chunhua Shen, Anthony Dick, and Anton Van Den~Hengel. 2017{\natexlab{a}}.
\newblock {FVQA}: Fact-based visual question answering.
\newblock \emph{TPAMI}, 40(10):2413--2427.

\bibitem[{Wang et~al.(2017{\natexlab{b}})Wang, Wu, Shen, Dick, and van~den Hengel}]{wangkbvqa}
Peng Wang, Qi~Wu, Chunhua Shen, Anthony~R. Dick, and Anton van~den Hengel. 2017{\natexlab{b}}.
\newblock Explicit knowledge-based reasoning for visual question answering.
\newblock In \emph{IJCAI}.

\bibitem[{Wei et~al.(2024)Wei, Chen, Chen, Hu, Zhang, Fu, Ritter, and Chen}]{wei2023uniir}
Cong Wei, Yang Chen, Haonan Chen, Hexiang Hu, Ge~Zhang, Jie Fu, Alan Ritter, and Wenhu Chen. 2024.
\newblock Uniir: Training and benchmarking universal multimodal information retrievers.
\newblock \emph{ECCV}.

\bibitem[{Weston et~al.(2015)Weston, Chopra, and Bordes}]{weston2015memory}
Jason Weston, Sumit Chopra, and Antoine Bordes. 2015.
\newblock Memory networks.
\newblock In \emph{ICLR}.

\bibitem[{Wolf et~al.(2020)Wolf, Debut, Sanh, Chaumond, Delangue, Moi, Cistac, Rault, Louf, Funtowicz, Davison, Shleifer, von Platen, Ma, Jernite, Plu, Xu, Scao, Gugger, Drame, Lhoest, and Rush}]{wolf-etal-2020-transformers}
Thomas Wolf, Lysandre Debut, Victor Sanh, Julien Chaumond, Clement Delangue, Anthony Moi, Pierric Cistac, Tim Rault, Rémi Louf, Morgan Funtowicz, Joe Davison, Sam Shleifer, Patrick von Platen, Clara Ma, Yacine Jernite, Julien Plu, Canwen Xu, Teven~Le Scao, Sylvain Gugger, Mariama Drame, Quentin Lhoest, and Alexander~M. Rush. 2020.
\newblock \href {https://www.aclweb.org/anthology/2020.emnlp-demos.6} {Transformers: State-of-the-art natural language processing}.
\newblock In \emph{Proceedings of the 2020 Conference on Empirical Methods in Natural Language Processing: System Demonstrations}, pages 38--45, Online. Association for Computational Linguistics.

\bibitem[{Workshop et~al.(2022)Workshop, Scao, Fan, Akiki, Pavlick, Ili{\'c}, Hesslow, Castagn{\'e}, Luccioni, Yvon et~al.}]{workshop2022bloom}
BigScience Workshop, Teven~Le Scao, Angela Fan, Christopher Akiki, Ellie Pavlick, Suzana Ili{\'c}, Daniel Hesslow, Roman Castagn{\'e}, Alexandra~Sasha Luccioni, Fran{\c{c}}ois Yvon, et~al. 2022.
\newblock Bloom: A 176b-parameter open-access multilingual language model.
\newblock \emph{arXiv preprint arXiv:2211.05100}.

\bibitem[{Wu et~al.(2016)Wu, Wang, Shen, Dick, and Van Den~Hengel}]{wu2016ask}
Qi~Wu, Peng Wang, Chunhua Shen, Anthony Dick, and Anton Van Den~Hengel. 2016.
\newblock Ask me anything: Free-form visual question answering based on knowledge from external sources.
\newblock In \emph{CVPR}.

\bibitem[{Xiao et~al.(2024)Xiao, Gong, Cascante-Bonilla, Zhang, Wu, and Ordonez}]{xiao2024grounding}
Zilin Xiao, Ming Gong, Paola Cascante-Bonilla, Xingyao Zhang, Jie Wu, and Vicente Ordonez. 2024.
\newblock Grounding language models for visual entity recognition.
\newblock \emph{arXiv preprint arXiv:2402.18695}.

\bibitem[{Yang et~al.(2022)Yang, Gan, Wang, Hu, Lu, Liu, and Wang}]{pica}
Zhengyuan Yang, Zhe Gan, Jianfeng Wang, Xiaowei Hu, Yumao Lu, Zicheng Liu, and Lijuan Wang. 2022.
\newblock An empirical study of gpt-3 for few-shot knowledge-based {VQA}.
\newblock In \emph{AAAI}.

\bibitem[{Ye et~al.(2023)Ye, Xu, Xu, Ye, Yan, Zhou, Wang, Hu, Shi, Shi et~al.}]{ye2023mplug}
Qinghao Ye, Haiyang Xu, Guohai Xu, Jiabo Ye, Ming Yan, Yiyang Zhou, Junyang Wang, Anwen Hu, Pengcheng Shi, Yaya Shi, et~al. 2023.
\newblock mplug-owl: Modularization empowers large language models with multimodality.
\newblock \emph{arXiv preprint arXiv:2304.14178}.

\bibitem[{Zhang et~al.(2022)Zhang, Roller, Goyal, Artetxe, Chen, Chen, Dewan, Diab, Li, Lin et~al.}]{zhang2022opt}
Susan Zhang, Stephen Roller, Naman Goyal, Mikel Artetxe, Moya Chen, Shuohui Chen, Christopher Dewan, Mona Diab, Xian Li, Xi~Victoria Lin, et~al. 2022.
\newblock Opt: Open pre-trained transformer language models.
\newblock \emph{arXiv preprint arXiv:2205.01068}.

\bibitem[{Zhang et~al.(2018)Zhang, Dai, Kozareva, Smola, and Song}]{zhang2018variational}
Yuyu Zhang, Hanjun Dai, Zornitsa Kozareva, Alexander Smola, and Le~Song. 2018.
\newblock Variational reasoning for question answering with knowledge graph.
\newblock In \emph{AAAI}.

\bibitem[{Zhou et~al.(2017)Zhou, Yao, Wen, Wang, Zhou, He, and Liang}]{zhou2017east}
Xinyu Zhou, Cong Yao, He~Wen, Yuzhi Wang, Shuchang Zhou, Weiran He, and Jiajun Liang. 2017.
\newblock East: an efficient and accurate scene text detector.
\newblock In \emph{CVPR}.

\bibitem[{Zhu et~al.(2023)Zhu, Chen, Shen, Li, and Elhoseiny}]{zhu2023minigpt}
Deyao Zhu, Jun Chen, Xiaoqian Shen, Xiang Li, and Mohamed Elhoseiny. 2023.
\newblock Minigpt-4: Enhancing vision-language understanding with advanced large language models.
\newblock \emph{arXiv preprint arXiv:2304.10592}.

\end{thebibliography}
\appendix
\section*{Appendix}
\section{Question Categorisation}
\label{sec:ques_categories}
We show the visual question-answering results over concretized sub-categories under each of the scenes, book and movie split in Table~\ref{tab:qa_resultsBreakDown}. We observe that our proposed model shows remarkable performance across diverse question categories, particularly in the challenging categories such as date, people, and open-ended question categories. 

\section{Finetuning details of LMMs}
In this section, we explain the hyperparameters and prompts used to finetune the LMMs. Note that we conduct all our experiments on a machine with 3 48GB A6000 GPUs. For mPlug-Owl and MiniGPT4v2, we have used hyperparameters as per the original papers.

\noindent\textbf{mPlug-Owl}: We finetuned mPlug-Owl with LoRA for 6 epochs with a learning rate of 2e-5 with a batch size of 256. LoRA details: rank: 8, alpha: 32, dropout: 0.05.
\begin{tcolorbox}[title=Instruction prompt template for mPlug-Owl] 
\footnotesize
The following is a conversation between a curious human and an AI assistant. The assistant gives accurate and crisp answers to the user's questions.
\\Human: \textcolor{blue}{<image>}
\\Human: \textcolor{blue}{\{Q\}}
\\AI: \textcolor{red}{\{A\}}.
\end{tcolorbox}

\noindent\textbf{MiniGPTv4v2}: We finetuned MiniGPTV4v2 with LoRA for 6 epochs with a learning rate of 3e-5 with a batch size of 128. LoRA details: rank: 16, alpha: 64, dropout: 0.05.
\begin{tcolorbox}[title=Instruction prompt template for MiniGPT4v2] 
\footnotesize
\textcolor{blue}{<image>}
\\ \{vqa\} Based on the image, respond to this question with a short answer: \textcolor{blue}{\{Q\}}, ASSISTANT: \textcolor{red}{\{A\}}
\end{tcolorbox}

\noindent\textbf{InstructBLIP}: We finetuned InstructBLIP for 3 epochs with a learning rate of 1e-5 with a batch size of 128.
\begin{tcolorbox}[title=Instruction prompt template for InstructBLIP] 
\footnotesize
\textcolor{blue}{<image>}
\\ USER: \textcolor{blue}{\{Q\}}. ASSISTANT: \textcolor{red}{\{A\}}
\end{tcolorbox}

\noindent\textbf{LLaVA-1.5}: We finetine LLaVA with LORA for 6 epochs with a learning rate of 5e-5 with a batch size of 64. LoRA details: rank: 16, alpha: 32, dropout: 0.05.

\begin{tcolorbox}[title=Instruction prompt template for LLaVA-1.5] 
\footnotesize
\textcolor{blue}{<image>}
\\ USER: \textcolor{blue}{\{Q\}}. ASSISTANT: \textcolor{red}{\{A\}}
\end{tcolorbox}

\section{More Results}
\label{sec:app_more_results}
\begin{figure*}[t!]
\centering
  \includegraphics[width=\textwidth]{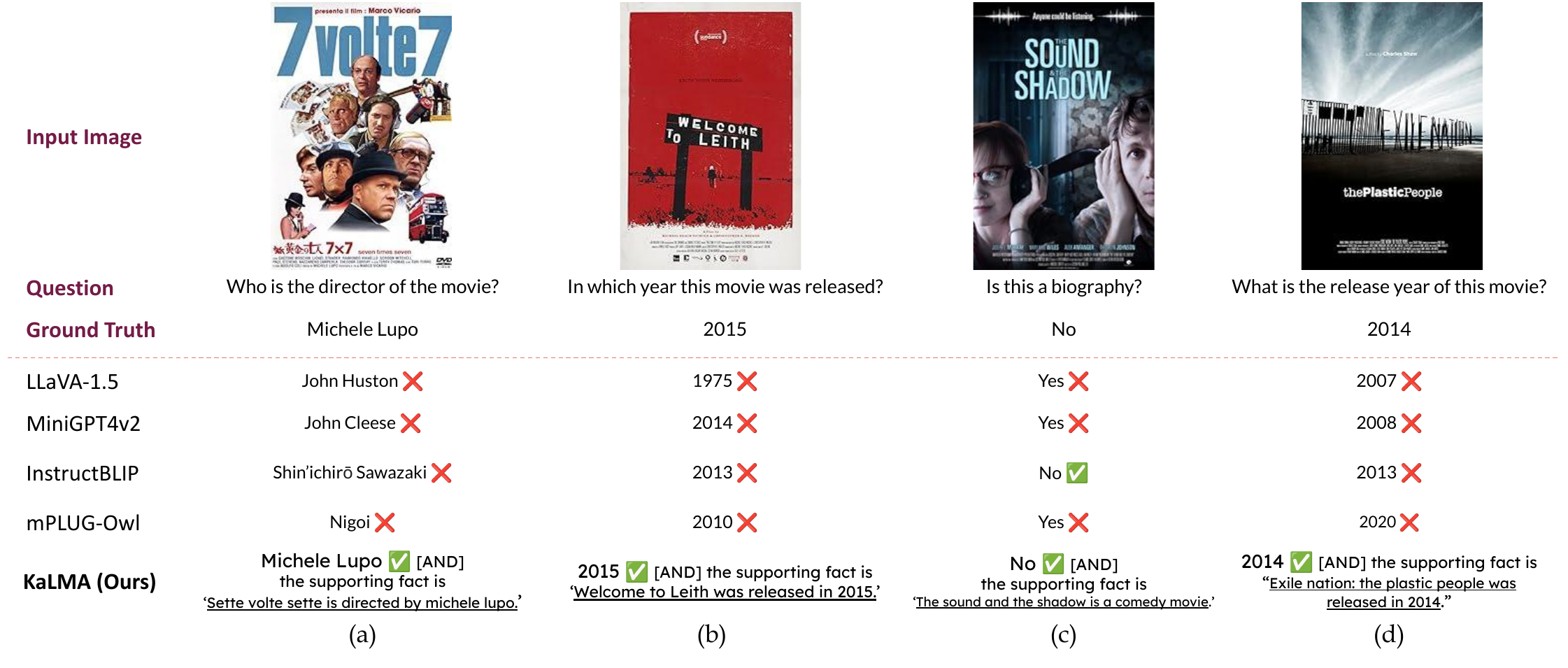}
  \caption{\textbf{A few more selection of our results as compared to implicit knowledge-based {\sc lmm} approaches on the movie subset of \data{}.}}
  \label{fig:select_results_movie}
\end{figure*}

\begin{figure*}[t!]
\centering
  \includegraphics[width=\textwidth]{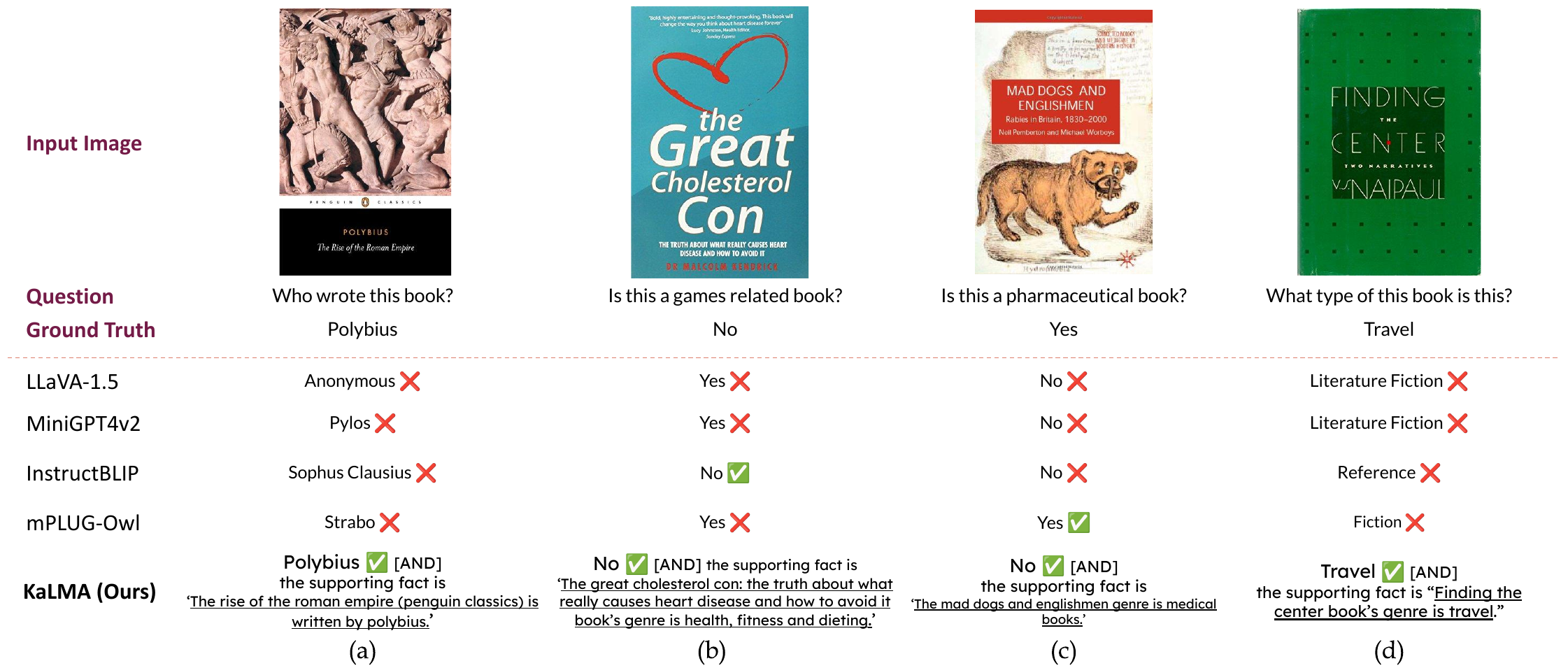}
  \caption{\textbf{A few more selection of our results as compared to implicit knowledge-based {\sc lmm} approaches on the book subset of \data{}.}}
  \label{fig:select_results_book}
\end{figure*}

More qualitative results on movie and book splits of \data{} are shown in Figure~\ref{fig:select_results_movie} and Figure~\ref{fig:select_results_book}, respectively. 

\begin{table*}[t]
    \centering
    \scriptsize
     \footnotesize
    \resizebox{\textwidth}{!}{
    \begin{tabular}{l ccccc cccccc ccccc}
\hline
    \multicolumn{1}{c}{} & \multicolumn{5}{c}{\data{} (scene)} & \multicolumn{5}{c}{\data{} (book)} & \multicolumn{6}{c}{\data{} (movie)} \\ 
    \cmidrule(r){2-6}
    \cmidrule(r){7-11}
    \cmidrule(r){12-17}
    Method & B & D & P & L & OE & B & D & P & G & OE & B & D & P & G & L & OE
    \\
\hline
    \textbf{Pre-LLM Methods} & & & & \\
    ~~~~GPT-2 & 54.8 & 0.2 & 0.0 & 13.7 & 15.4 & 54.5 & 43.8 & 0.1 & 4.3 & 0.6 & 74.5 & 2.1 & 0.0 & 15.2 & 63.7 & 0.0 \\
    ~~~~GPT-2 (w/ Visual Context) & 57.1 & 0.3 & 0.0 & 16.1 & 17.0 & 80.1 & 63.8 & 5.2 & 45.1 & 7.5 & 75.4 & 3.2 & 0.0 & 24.3 & 66.8 & 29.3 \\
    ~~~~ViLT & 75.9 & 0.0 & 0.0 & 33.9 & 28.7 & 68 & 63.3 & 0 & 21.3 & 0.9 & 85 & 4.4 & 0.2 & 42.1 & 76.7 & 0.0\\
    ~~~~VLBart & 78.9 & 0.2 & 0.0 & 18.8 & 27.4 & 79.2 & 62.0 & 1.7 & 34.9 & 0.9 & \textbf{85.4} & 6.3 & 0.0 & 43.7 & 76.7 & 0.0\\
    \hline
    \textbf{LLM Methods} \\
    ~~~~mPlug-Owl & 22 & 8.9 & 0.0 & 45 & 9.8 & 19.5 & 69.7 & 38.7 & 43.8 & 12 & 7.8 & 17.5 & 0.7 & 9.7 & 6.2 & 5.5 \\
    ~~~~LLaVA-1.5 & 81.1 & 0.0 & 2.0 & 38.7 & 23.4 & 79 & 70.6 & 19.3 & 57.3 & 2.7 & 84.8 & 13.5 & 0.3 & 1.6 & 72.7 & 9.9 \\
    ~~~~MiniGPT4v2 & \textbf{81.7} & 2.7 & 1.3 & 49.9 & 41.7 & 80.1 & 71.9 & 18.2 & 54.2 & 6.6 & 79.9 & 13.6 & 1.2 & 53.7 & 78.4 & 30.4\\ 
    ~~~~InstructBLIP & 50.0 & 0.1 & 6.6 & 29.7 & 32.8 & 49.8 & 70.3 & 22 & 15.2 & 12.8 & 50.0 & 6.6 & 0.3 & 1.4 & 76.5 & 39.5 \\
    \hline
    \textbf{Ours} \\
    ~~~~\textbf{\model{}} & 77.2 & \textbf{69.0} & \textbf{76.8} & \textbf{67.8} & \textbf{69.9} & \textbf{88.5} & \textbf{72.9} & \textbf{80.0} & \textbf{80.2} & \textbf{79.6} & 84.2 & \textbf{69.6} & \textbf{74.8} & \textbf{70.6} & \textbf{91.5} & \textbf{69.1}\\
    ~~~~\textbf{\model{} (Oracle)} & 83.9 & 95.8 & 95.4 & 91.9 & 91.8 & 98.0 & 96.4 & 98.2 & 99.9 & 98.2 & 99.9 & 99.8 & 95.9 & 100.0 & 100.0 & 99.7\\
 \hline
  \end{tabular}}
    \caption{QA accuracy performance breakdown for various methods by question categories on \textsc{\data{}}. Categories are \textbf{B}: binary, \textbf{D}: date, \textbf{P}: people, \textbf{L}: location, \textbf{G}: genre and \textbf{OE}: open-ended.}
  \label{tab:qa_resultsBreakDown}
\end{table*}

%


\end{document}